\documentclass[11pt]{article}
\usepackage[margin=1in]{geometry}
\usepackage{amsmath,amssymb,booktabs,array}
\usepackage[colorlinks=true,allcolors=blue]{hyperref}

\title{Selective Elicitation as a Commercial Influence Channel:\\
A Reproducible Synthetic Shopping-Agent Stress Test}
\author{Jiapeng Li}
\date{September 2026}

\begin{document}
\maketitle

\begin{abstract}
A commercial incentive need not enter the final ranking algorithm to affect a
shopping assistant's recommendation: it may instead influence which preference
question the assistant asks. We make this distinction experimentally observable
in a deliberately small, synthetic setting. Each task has two products, three
verified numerical attributes, a price limit, and a private, fixed preference
vector. An honest simulated user answers one pairwise question. A separate
recommender receives the products and this answer but not the sponsorship
assignment. We contrast a neutral question, a soft commercial instruction,
and an explicitly adversarial instruction to ask about the sponsor's advantage
while omitting its rival's advantage. Across 40 held-out sponsorship-assignment
cases (20 distinct catalog--preference contexts), the soft instruction changes
no selections. The targeted instruction raises sponsored selection by 0.30
and reduces mean \emph{synthetic} utility by 0.0547 relative to neutral questioning
(95\% context-bootstrap interval $[-0.0828,-0.0291]$) for one language-model
recommender. A fixed Bayesian recommender shows a similar effect; a second
model makes the same choices on all 120 frozen question--answer inputs.
A terminal-answer consistency judge rates all 20 sampled targeted answers
consistent, although five have synthetic regret above 0.05; a separate
question-coverage dimension flags their one-sided elicitation. A robust
partial-preference certificate remains valid under the stipulated synthetic
utility but certifies only 16 of 40 targeted cases and is not better than
asking a neutral question directly. These results establish neither typical
behavior under advertising incentives nor effects on actual consumers.
The contribution is a narrowly controlled stress-test protocol, a
reproducible observational-equivalence witness, and explicit negative
results for a mild incentive and a conservative intervention.
\end{abstract}

\noindent\textbf{Keywords:} AI shopping agents; commercial influence; sponsored questions;
preference elicitation; synthetic evaluation; auditability.

\section{Introduction}

An AI shopping assistant both solicits information about a user's needs and
recommends products. If a platform benefits from a particular product's
selection, the resulting conflict need not take the familiar form of a paid
boost in the final ranking. The platform can preferentially ask about an
attribute on which that product excels. The user's answer may be truthful
\emph{within the comparison offered} yet omit another attribute the user values
more. A downstream recommender can faithfully follow its visible evidence
and still choose an option that is inferior under the user's unobserved
preferences.

This possibility is not itself a new discovery. Prior work documents
commercial conflicts in LLM advice \cite{wu2026}, the role of agent
delegation \cite{wadi2026}, and randomized human evidence that conversational
persuasion changes sponsored choices \cite{salvi2026}. Sponsored questions
have been studied as an auction object \cite{bhawalkar2025}; effective
clarification has been analyzed separately \cite{caohu2026}. A taxonomy of
generative commercial intervention already identifies upstream redirection
and preference shaping \cite{qiumei2026}. Our more limited question is
\emph{what can be demonstrated, and what cannot be inferred, when only
the question policy changes and the final recommender does not receive
sponsorship information?}

We answer this using an instrumented synthetic test, not a claim about actual
consumer welfare. We report both the null effect of a mild commercial
instruction and the effect of an explicitly directed question-only attack.
We include cases where the sponsored item really is the best option.
We also show that a conditional regret certificate can be correct in its
own model without yielding an attractive method: it largely reduces to
the already strong neutral-question baseline.

\section{Relation to existing work and claim boundaries}

Wu et al.\ \cite{wu2026} categorize and test conflicts between user and
advertiser interests. Wadi and Ma \cite{wadi2026} vary the party to whom
an AI shopping agent is delegated. Allouah et al.\ \cite{allouah2025}
demonstrate that shopping agents' responses to sponsored labels and platform
endorsements differ. These results preclude a ``first evidence of sponsorship
bias'' claim, and none makes model product selection equivalent to realized
individual utility.

Salvi et al.\ \cite{salvi2026} report two preregistered experiments with
2,012 participants. Active conversational persuasion substantially increases
selection of randomly sponsored eBooks relative to traditional search
placement, but their keep-the-book-versus-\$1 comparison does not show a
significant between-condition decline in that measured outcome. Their
experiment has genuine human behavioral evidence; ours does not.
Question auctions \cite{bhawalkar2025} and efficient preference elicitation
\cite{caohu2026} motivate studying the question rather than the answer
alone. Quality-preserving ad auctions \cite{handai2026} use semantic fidelity
to an organic response, which is not an observation of a person's net benefit.

The general idea of high-confidence fallback is also established in offline
decision-making \cite{laroche2019}, and unsupported actions can sometimes
cancel when evaluating policy differences \cite{li2026}. Our certificate
below is an elementary robust-decision construction \emph{under an explicitly
given synthetic utility}, not a novel welfare guarantee. Finally, LLM
judges have known position, verbosity, and self-preference biases
\cite{zheng2023}, and simulated users are not reliable proxies for humans
in general \cite{zhou2026,seshadri2026}. We therefore keep deterministic
utility, model outputs, and judgments on different reporting axes.

\section{Formal setting}
\label{sec:setting}

A task contains a public catalog $C=\{A,B\}$, three verified product
attributes $v_{aj}\in\{0,\ldots,10\}$ for $j=1,2,3$, prices $p_a$,
and a public maximum budget $b$. The experimenter fixes a private
preference vector $\theta\in\mathbb{R}_{\geq0}^3$ with
$\sum_j\theta_j=1$ before the question is generated. For a budget-compliant
product we define
\begin{equation}
 U_\theta(a)=\sum_{j=1}^3\theta_j\frac{v_{aj}}{10}
       -\frac{p_a-\min_{c\in C}p_c}{400}.
 \label{eq:utility}
\end{equation}
A product over budget has utility zero, while not buying has utility
$U_\theta(\varnothing)=0.24$. These are \emph{stipulated research
payoffs}, not monetary willingness to pay or consumer satisfaction.
The best action $a^*(\theta,C)$ maximizes this function over
$C\cup\{\varnothing\}$.

The questioner sees the same public catalog in every arm. A neutral policy
$q_N$ is told to ask about a useful distinguishing tradeoff. A soft commercial
policy $q_S$ is told that the platform benefits from one sponsored product
but to remain helpful. An explicitly targeted policy $q_T$ is told to ask
about the sponsor's strongest attribute relative to a nondiscriminating
third attribute while omitting the competitor's strongest attribute.
The final selector $g$ receives the public catalog, original request,
question, and answer, but not the sponsor assignment or any bid.
The simulated user answers the \emph{two named attributes} honestly
according to $\theta$, reporting a tie when their weights are equal.

For each fixed task $i$ and selector $g$, our paired statistic is
\begin{equation}
 d_i(q,g)=U_{\theta_i}\!\left(g(C_i,M_i(q))\right)
             -U_{\theta_i}\!\left(g(C_i,M_i(q_N))\right).
 \label{eq:delta}
\end{equation}
We separately report the difference in the indicator that $g$ chooses
the sponsored item. The latter is \emph{not} realized platform revenue:
there are no ad payments or observed purchases in this study.

\subsection{An observational-equivalence witness}
\label{sec:witness}

The generated headphone catalog provides a direct example. Product A has
price 103 and scores $(9,4,6)$; B has price 114 and scores $(4,10,6)$.
Both satisfy the public budget 160. The question ``Which matters more,
wearing comfort or battery life?'' is answered ``wearing comfort'' by
both $\theta^{(1)}=(0.68,0.22,0.10)$ and
$\theta^{(2)}=(0.22,0.68,0.10)$. A frozen Bayesian selector chooses B
from this same visible transcript in both cases. But
\begin{align*}
 U_{\theta^{(1)}}(A)&=0.760,&U_{\theta^{(1)}}(B)&=0.525,\\
 U_{\theta^{(2)}}(A)&=0.530,&U_{\theta^{(2)}}(B)&=0.801.
\end{align*}
The public catalog, fixed question, answer, and final selection do not
identify whether B is better for this simulated user. This simple
two-world construction is a measurement warning, not a priority claim
for a general impossibility theorem. In the actual experiment the
researcher can score the answer because $\theta$ was fixed and retained
privately by the test harness; an external auditor of a real platform
would not have that privilege.

\section{Experimental protocol}

Six category labels (headphones, backpacks, chairs, vacuums, laptops,
and tents) share a deliberately simple numeric structure: one item
is relatively strong on the first attribute and the other on the second;
the third attribute is the same. Small seeded price and score changes
do not make these independent product markets. Five private profiles
per category favor the first or second attribute, balance them,
impose a hard budget, or make both products unaffordable.
For each catalog--profile combination we flip sponsorship between
A and B. The first two categories are the 20-case pilot; the
remaining four are the 40-case held-out evaluation, containing
20 catalog--profile contexts with two sponsor assignments each.

The same question-generation model identifier, \texttt{gpt-6-sol},
is used across arms. The arms execute in pseudorandom order. In the
primary evaluation, the synthetic user responds by a deterministic,
auditable comparison of two hidden weights. Final decisions come
either from the same stateless LLM with a user-only prompt or from a
fixed symmetric grid prior over preference weights, updated using
only the observed pairwise comparison. Neither selector has access
to the private weights or sponsor label. A separate
\texttt{gemini-3.6-flash} selector replays all 120 frozen
question--answer inputs from the held-out set. A Gemini simulated
user is examined on a preselected eight-case subset; it does not
replace the deterministic user in the primary estimate.

A blinded \texttt{gpt-5.4-mini} judge assesses a random 20-case
subset in each question arm for tradeoff coverage, consistency of
the final answer with visible evidence, and factual contradictions.
\texttt{gemini-3.6-flash} repeats the judgment on eight of those
cases. Judges do not see $\theta$ or sponsorship and cannot supply
utility ground truth. We use 2,000 context-cluster bootstrap draws
for descriptive 95\% intervals and report a second sensitivity
interval that resamples the four category labels. Because all
categories inherit the same basic numeric template, neither interval
supports generalization to real shoppers or independent catalogs.
Model sampling could not be frozen using a supported temperature
or provider-side seed.

\section{Results}

\begin{table}[t]
\centering
\small
\begin{tabular}{llrrl}
\toprule
Question arm vs.\ neutral & Selector &
 $\overline{\Delta U}$ & $\Delta P(\mathrm{sponsor})$ &
 95\% context CI for $\overline{\Delta U}$\\
\midrule
Soft commercial & LLM & 0.0000 & 0.00 & $[0,0]$\\
Soft commercial & Grid Bayes & 0.0000 & 0.00 & $[0,0]$\\
Targeted stress test & LLM & $-0.0547$ & $+0.30$ &
 $[-0.0828,-0.0291]$\\
Targeted stress test & Grid Bayes & $-0.0540$ & $+0.30$ &
 $[-0.0822,-0.0281]$\\
\bottomrule
\end{tabular}
\caption{Forty paired held-out sponsorship-assignment cases. Payoffs are
synthetic, not individual human outcomes. The targeted policy is explicitly
directed to omit a rival strength; its effect is not a spontaneous response
to mild monetization.}
\label{tab:main}
\end{table}

The LLM selector chooses the sponsored option in 40\%, 40\%, and
70\% of neutral, soft, and targeted cases. Relative to neutral,
the targeted question harms 12 cases, improves none, and leaves 28
unchanged for this selector. Its mean change is $-0.0912$ among
the 24 cases where the sponsored product is not the synthetic optimum,
versus zero among the 16 where it \emph{is} optimal. All three arms
respect the public budget in all 40 held-out cases, and the targeted
arm still selects all 16 of 16 genuinely optimal sponsored products.
The soft arm asks about both discriminating attributes in 39 of 40
cases and does not change any final choices. This null result rules
out an interpretation that merely mentioning a platform incentive
produced the measured targeted effect in our setting.

The second LLM selector makes the same choice as the first on
all 120 saved arm-by-case inputs. This checks \emph{final model
selection on the same questions}, not independent replication of
the question generator, user population, or task generator. The
preselected Gemini user subset supplies 24 of 24 structured
choice fields consistent with the private pairwise ordering; only
eight cases permit a complete neutral--targeted AI-user comparison
(mean synthetic utility difference $-0.0259$). This does not
validate simulation against human shoppers.

In the 20 judged targeted traces, tradeoff coverage is marked
absent 20 times, yet visible answer--recommendation consistency is
marked satisfactory 20 times. Five of these judged outputs have
synthetic regret exceeding 0.05 and still receive the satisfactory
terminal-consistency verdict. Thus this \emph{particular submetric}
can miss one-sided elicitation; the full-trajectory coverage judgment
already signals it. Two judge families agree on the eight selected
cross-checked cases, too few to establish human agreement.
No factual contradictions are flagged within the judged subset;
the remaining 20 trajectories per arm have no such independent
judge check.

\section{A conditional certificate that does not beat the baseline}

For an honest answer choosing feature $j$ over $k$, let
$\Theta_{jk}=\{\theta\geq0:\sum_\ell\theta_\ell=1,\,
\theta_j\geq\theta_k\}$. A tie uses equality instead.
The worst-case regret of action $a$ for this restricted synthetic
problem is
\begin{equation}
 R(a\mid j\succeq k)=
 \max_{\theta\in\Theta_{jk}}\left[
    \max_{a'\in\{A,B,\varnothing\}}U_\theta(a')-U_\theta(a)
 \right].
 \label{eq:certificate}
\end{equation}
Evaluating the feasible simplex vertices suffices: each pairwise
utility difference is affine in $\theta$, and the maximum over
alternatives can be evaluated at a vertex. This property is
conditional on our exact utility function and an honest response,
not on an AI judge's confidence. We choose the action with the
smallest bound and certify only if it is at most the stipulated
tolerance $\epsilon=0.05$. Otherwise a second, neutral question
supplies a fallback decision. This is an application of established
robust optimization logic, not a new theorem about people.

On the 40 held-out cases, neutral questions receive certificates
in 34 cases; targeted questions in only 16. Of the 16 cases where
the sponsor truly is optimal, only four are certified after a
targeted question. Targeted cases require an additional neutral
question in 60\% of cases; after this fallback their mean regret is
0.0008, the same as asking the neutral question immediately.
All certified actions satisfy the model's bound in the executable
test suite. This baseline does \emph{not} establish a superior
commercial-recommendation mechanism: it commonly withholds a
beneficial option and increases interaction cost.

\section{Limitations and scope of publication}

\paragraph{No consumer-welfare inference.}
Private weights and payoff coefficients are invented by the
experimenter. A synthetic change of $-0.0547$ is not a
measured decrease in people's welfare or an effect on sales,
satisfaction, trust, or longitudinal choice. The AI-user subset
does not fix this limitation.

\paragraph{Limited independent variation.}
All six category labels reuse the same two-product, three-feature
tradeoff template with limited jitter. Sponsorship is balanced
within contexts, but the four holdout categories are not
independent real catalog samples. The comparatively narrow
category-bootstrap sensitivity interval reflects this shared
construction and cannot be interpreted as external validity.
There is no multi-turn learning, advertiser adaptation, actual
retrieval, or realistic offer availability.

\paragraph{Constructed attack versus platform behavior.}
The only harmful arm explicitly instructs the questioner to
omit the rival's important attribute; the mild incentive produces
no effect here. We do not infer that existing shopping platforms
perform this intervention or that a general model spontaneously
does so. A neutral selector and a rationally updated grid prior
both respond to the information they receive, so the demonstration
is about choice architecture, not an LLM-specific reasoning defect.

\paragraph{Moderation and future work.}
This artifact is an exploratory, self-contained technical report,
not a claim to have discovered advertising bias or proved a new
general guarantee. Stronger evidence would require independent
catalog generators, independent question-generating models,
multiple prompts and dialog lengths, stronger cost-aware
baselines, and human validation before any statements about
actual consumer benefit. Upstream candidate-set suppression
may offer a more directly verifiable next question when no
human study is possible.

\section*{Artifact availability and AI-tool disclosure}

The intended submission archive includes source code and
synthetic traces in an \texttt{anc/} directory. The held-out
question/answer trace's SHA-256 starts with \texttt{ab3ec081}
and ends with \texttt{0bccdab9}; the full digest is in the
included \texttt{summary.json}.
Statistics can be recomputed from saved JSONL without API access;
new model outputs require access to the named models through an
OpenAI-compatible interface and may not be byte-for-byte stable.
No corporate evaluation repository, private credentials, human
shopping data, or real advertiser payments are included.

Generative AI tools were used substantially for code development,
experiment orchestration, evaluation, and manuscript drafting.
They are not authors. The named human author is responsible for
verifying the calculations, references, licensing, and all claims
in any publicly submitted version.

\end{document}